\documentclass[runningheads]{llncs}
\usepackage{graphicx}
\usepackage{multirow}
\usepackage{multicol}
\usepackage{comment}
\usepackage{amsmath,amssymb} 
\usepackage{color}
\usepackage{wrapfig}

\begin{document}
\pagestyle{headings}
\mainmatter
\def\ECCVSubNumber{15}  

\title{GIA-Net: Global Information Aware Network for Low-light Imaging} 

\titlerunning{GIA-Net: Global Information Aware Network for Low-light Imaging}
%
\author{Zibo Meng\inst{1}\orcidID{https://orcid.org/0000-0001-7299-7290} \and
Runsheng Xu\inst{2} \and
Chiu Man Ho\inst{1}}
\authorrunning{Z. Meng et al.}
%
\institute{InnoPeak Technology, Palo Alto CA 94043 
\email{\{zibo.meng,chiuman\}@innopeaktech.com}
\and
Mercedes-Benz R\&D North America, Sunnyvale CA 94085\\
\email{derrickxu1994@gmail.com}}
\maketitle

\begin{abstract}
It is extremely challenging to acquire perceptually plausible images under low-light conditions due to low SNR. Most recently, U-Nets have shown promising results for low-light imaging. However, vanilla U-Nets generate images with artifacts such as color inconsistency due to the lack of global color information. In this paper, we propose a global information aware (GIA) module, which is capable of extracting and integrating the global information into the network to improve the performance of low-light imaging. The GIA module can be inserted into a vanilla U-Net with negligible extra learnable parameters or computational cost. Moreover, a GIA-Net is constructed, trained and evaluated on a large scale real-world low-light imaging dataset. Experimental results show that the proposed GIA-Net outperforms the state-of-the-art methods in terms of four metrics, including deep metrics that measure perceptual similarities. Extensive ablation studies have been conducted to verify the effectiveness of the proposed GIA-Net for low-light imaging by utilizing global information.
\end{abstract}

\section{Introduction}

Taking photos with good perceptual quality under low illumination conditions is extremely challenging due to low signal-to-noise ratio (SNR)~\cite{chen2018learning}. One common practice to improve the low-light image quality is to extend the exposure time.  However, this can easily introduce motion blur due to camera shake or object movements and it is not always applicable in real life. In the past decade, extensive studies have been conducted for imaging under low-light conditions including denoising techniques~\cite{rudin1992nonlinear,portilla2003image,mairal2009non,gu2014weighted,dabov2007image,jain2009natural,zhang2017beyond} which aim at removing  noises introduced in the acquired low-light images, and image enhancement techniques~\cite{dong2011fast,malm2007adaptive,loza2013automatic,park2017low,guo2017lime} which are developed for improving the perceptual quality of digital images. 
\begin{figure}[]
\centering
\includegraphics[width=0.7\textwidth]{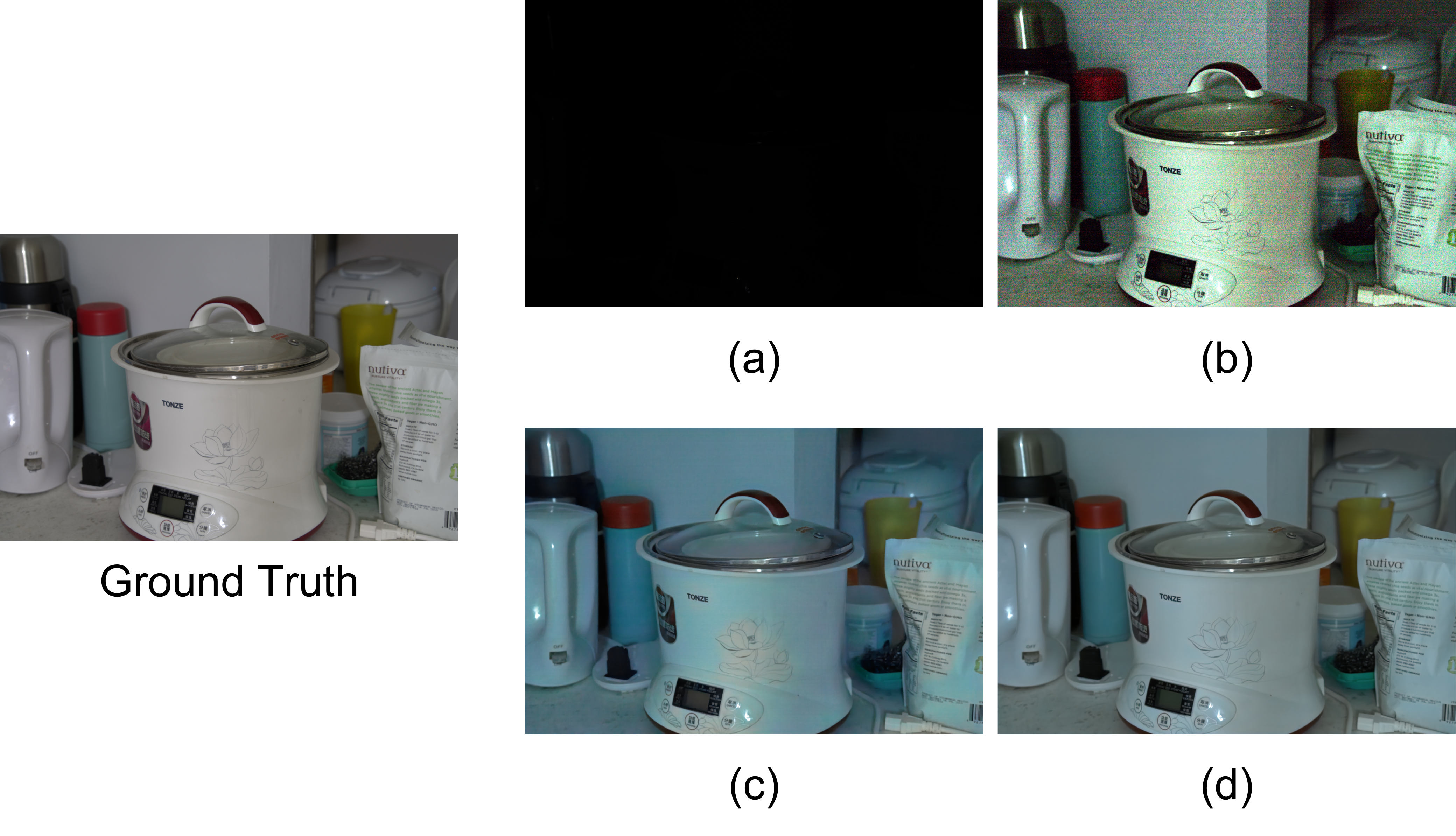}
\caption{An illustration of low-light imaging. (a) A short-exposed RAW input taken from the SONY subset of the SID dataset~\cite{chen2018learning} with an exposure time of 0.04s; (b) The RGB image produced by applying traditional image signal processing pipeline to the short exposed raw image given in (a). Note that the brightness has been increased for better representation; (c) The RGB image produced using the state-of-the-art approach~\cite{chen2018learning}. Note that severe artifacts, such as color inconsistency, can be spotted in the resulting image; (d) The output of the proposed GIA-Net, where the color of the image is consistent everywhere because of the introduction of global information.}
\label{fig:comparison}
\end{figure}

Most recently, deep convolutional neural networks~\cite{chen2018learning,zamir2019learning} have shown promise for imaging under low-light conditions. Specifically, Chen \textit{et al.} developed a framework based on a U-Net using $\ell_1$ loss function as the objective function. Following Chen's work, Zamir \textit{et al.}~\cite{zamir2019learning} proposed a new joint loss function to train the U-Net with the same architecture as in~\cite{chen2018learning} for low-light imaging. Although inspiring results have been presented in those work, both of the proposed methods produced severe artifacts, such as color inconsistency, due to the lack of global information in the network. For example, as illustrated in Figure~\ref{fig:comparison}, Fig.~\ref{fig:comparison}(a) gives a short-exposed RAW input taken from the SONY subset of the SID dataset~\cite{chen2018learning} with an exposure time of 0.04s; Fig.~\ref{fig:comparison} (b) depicts the output image produced by applying traditional digital signal processing pipeline to the short exposed image (a). Note the high noise level and color distortion; Fig.~\ref{fig:comparison} (c) shows the output image produced using the state-of-the-art approach~\cite{chen2018learning}. Note the color inconsistency in the output image because of the lack of the global information in the U-Net employed.

To overcome the shortcomings of the vanilla U-Nets for low-light imaging, in this work, we develop a framework for imaging under extremely low-light conditions in an end-to-end fashion with global color information integrated. Specifically, we propose a global information-aware (GIA) module for low-light imaging, which is capable of extracting global information, together with the pixel-level features, to improve the perceptual qualities for low-light image enhancement. Furthermore, we insert the proposed GIA module into a vanilla U-Net to construct a GIA-Net. As illustrated in Fig.~\ref{fig:comparison} (d), the output of our GIA-Net gives consistent color compared with Fig.~\ref{fig:comparison} (c) produced by~\cite{chen2018learning}.  The GIA-Net can be trained in an end-to-end fashion with a joint loss function. \textbf{ The code for training and testing, as well as the trained models will be publicly available}.

Our main contributions are threefold:

We propose a GIA module to extract and integrate global information into U-Nets;

We design a GIA-Net with the proposed GIA module integrated, and demonstrated its effectiveness for low-light imaging;

We conduct extensive ablation study to demonstrate the effectiveness of the proposed GIA-Net for low-light imaging.


\section{Related Work}

Image processing and enhancement have been extensively studied in the past decades which are discussed in the following sections.

\subsection{Image denoising}
Image denoising has been widely studied in low-level vision field. 

\textbf{Single image denoising}, such as total variation denoising~\cite{rudin1992nonlinear} and 3D transform-domain filtering (BM3D)~\cite{dabov2007image} for image denoising, is often based on analytical priors such as image smoothness, sparsity, low rank, or self-similarity to recover the image signals from noisy images. In the past few years, because of their extraordinary performance in other computer vision applications, deep convolutional neural networks (CNNs) have been emerging for image denoising~\cite{zhang2017beyond,zhang2018ffdnet,zhang2017beyond,anwar2019real}. While remarkable improvement has been achieved, those methods are generally developed and evaluated on synthetic data and do not generalize well to real images. Most recently, while elf-guided network~\cite{gu2019self,kneubuehler2020flexible} has been proposed and shown promise for image enhancement, its performance might degrade for low-light images since it directly uses the highly noisy images as input for every level.

\textbf{Burst denoising} performs denoising on burst of noisy images captured sequentially using the same device from the same scene~\cite{hasinoff2016burst,liu2014fast}. Those approaches generally first register all the frames to a common reference, and then perform denoising by robust averaging~\cite{mildenhall2018burst}.
In addition, a set of approaches is using a burst of images taken at the same time to perform denoising. Although these methods typically yield good performance, they are elaborately and computationally expensive. Moreover, image alignment algorithms become unreliable under extremely low-light conditions, resulting in ghosting effects in the final image. 

\subsection{Low-light image enhancement}
A number of techniques have been developed for image enhancement, such as histogram equalization, and gamma correction. Recently, more advanced approaches have been proposed to deal with the enhancement of low-light images\cite{dong2011fast,malm2007adaptive,loza2013automatic,park2017low,guo2017lime,gharbi2017deep,chen2018deep} . However, these models share a strong assumption where the input image has clean representation without any noise. Thus, a separate denoising step should be employed beforehand for low-light image enhancement.
One particular method that is related to our approach is the ``learning to see in the dark'' model (SID)~\cite{chen2018learning} where an encoder-decoder CNN is employed to perform denoising, as well as image enhancement at the same time. In a follow-up work~\cite{zamir2019learning}, a joint loss function, i.e. $\ell_1$, MS-SSIM~\cite{wang2003multiscale}, and perceptual loss~\cite{johnson2016perceptual}, is proposed to improve the quality of the generated images. However, since the global information is not considered in both of the work, severe artifacts such as color inconsistency can be observed in the output images.

Most of the current approaches perform image denoising and enhancement separately, which is time and computationally costly. Moreover, although SID~\cite{chen2018learning} performed image denoising and enhancement jointly and achieved promising results, it failed to consider the global information which is crucial for color consistency in the output images. In this work, we propose to perform low-light image denoising and enhancement in a single shot with the integration of the global context. This  makes the network to be aware of the global context/color information to better generate the final output.

\section{Methodology}

In this section, we firstly present some analyses on drawbacks of applying vanilla U-Nets on low-light imaging as proposed in ~\cite{chen2018learning,zamir2019learning}. Then, we introduce a global information aware (GIA) module to deal with the drawbacks and insert the proposed GIA module into a U-Net for low-light imaging.

\subsection{Analysis on vanilla U-Nets}
U-Nets have been widely adopted for image-to-image translation and have been demonstrated to be effective for semantic segmentation. However, vanilla U-Nets have some drawbacks for low-light imaging. For example, 
as illustrated in Fig.~\ref{fig:comparison} (c), color inconsistency can be observed in the generated result using a vanilla U-Net~\cite{chen2018learning} due to the lack of global color information. Specifically, the effective receptive size of the network used in~\cite{chen2018learning} is around 224,  while the input image of the network is $2832\times 4240$ for images in Sony dataset. Thus, we develope a global information aware (GIA) module which can be inserted into a U-Net to extract and utilize the global information for low-light imaging.

\begin{figure}
\centering
\includegraphics[width=0.5\textwidth]{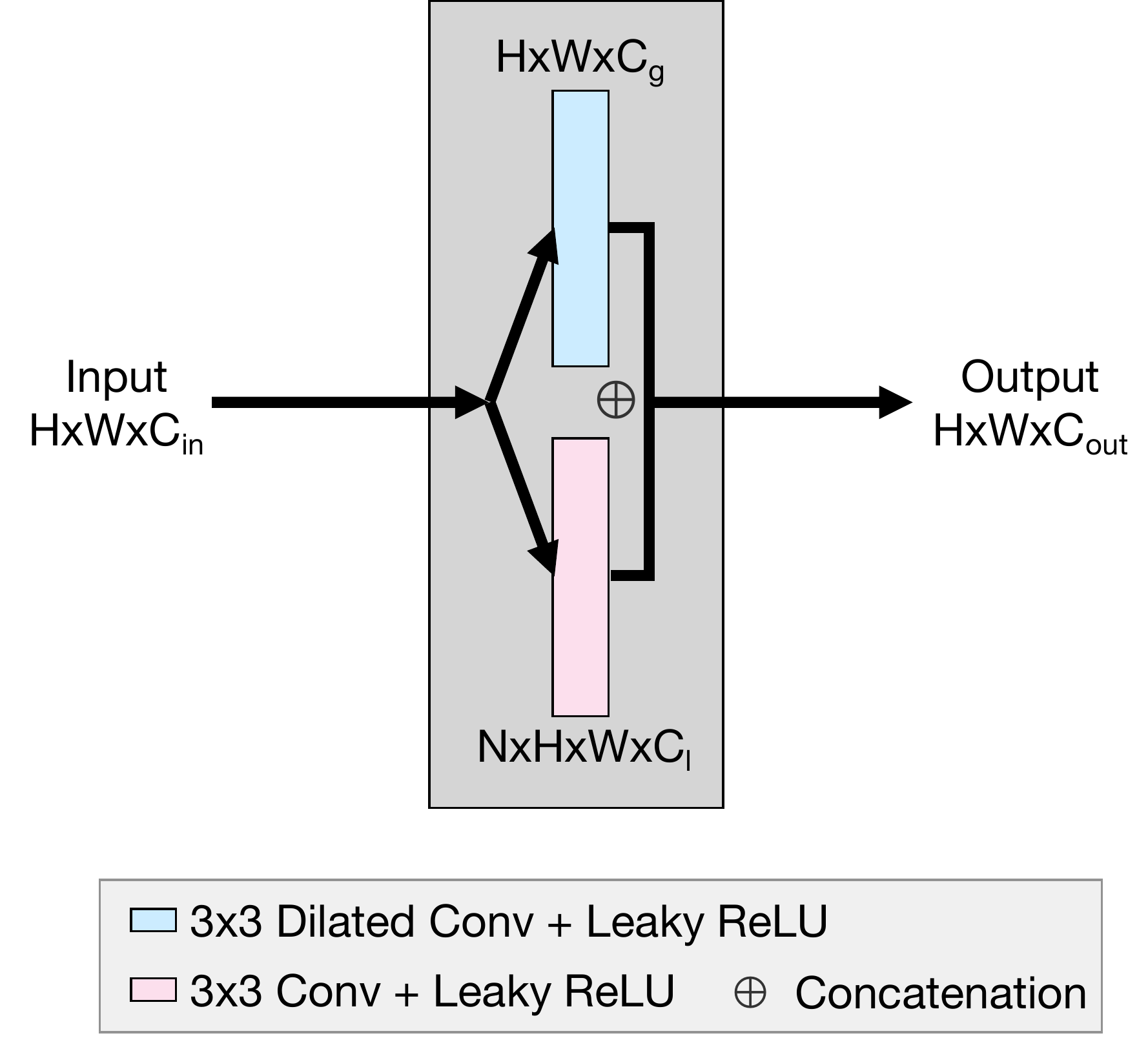}
\caption{An illustration of the seeing wider (SW) block, where the input will be fed into a vanilla convoluitonal layer, which is responsible to extract the local information, and a dilated convolutional layer, which is responsible to extract contextual information in a wider range. The outputs are concatenated as the final output of the SW block.}
\label{fig:seewider}
\end{figure}

\subsection{Global Information Aware Module}


The above-mentioned analysis motivates us to design a global information extraction module, i.e. GIA module, to extract and include the global information into the network to enable better performance for low-light imaging.

One natural choice is dilated convolutonal operation, which is widely adopted in deep convolutional neural networks for expanding the receptive field size. However, dilated convolutional operations ignore the local information and do not fully use all the information in the neighborhood. An alternative design is to use a combination of dilated convolutional operation and the vanilla convolutional operation. Specifically, as illustrated in Fig.~\ref{fig:seewider}, we design a see-wider (SW) module to enable the network to see both local fine details and wider contextual information. In a SW block, the input with a shape of $H\times W\times C_{in}$ will be separately fed into a vanilla convolutional layer with an output of a shape of $H\times W\times C_{l}$ and a dilated convolutional layer with an output of a shape of $H\times W\times C_{g}$. The outputs of the two layers are concatenated as the output of the SW block. Although the proposed SW block can integrate the local information with information extracted with a larger receptive field size, the design has three potential problems. First, the size of the input of the network can be arbitrarily large, while the receptive field size is fixed once the network is designed and trained. Second, another hyper-parameter, i.e. the dilate rate, is introduced, which needs extra effort to tune to achieve optimal performance. Third,  the numbers of dilated convolutional kernel and the regular convolutional kernel require to be determined through extensive experimental search.

\begin{figure}
\centering
\includegraphics[width=0.3\textwidth]{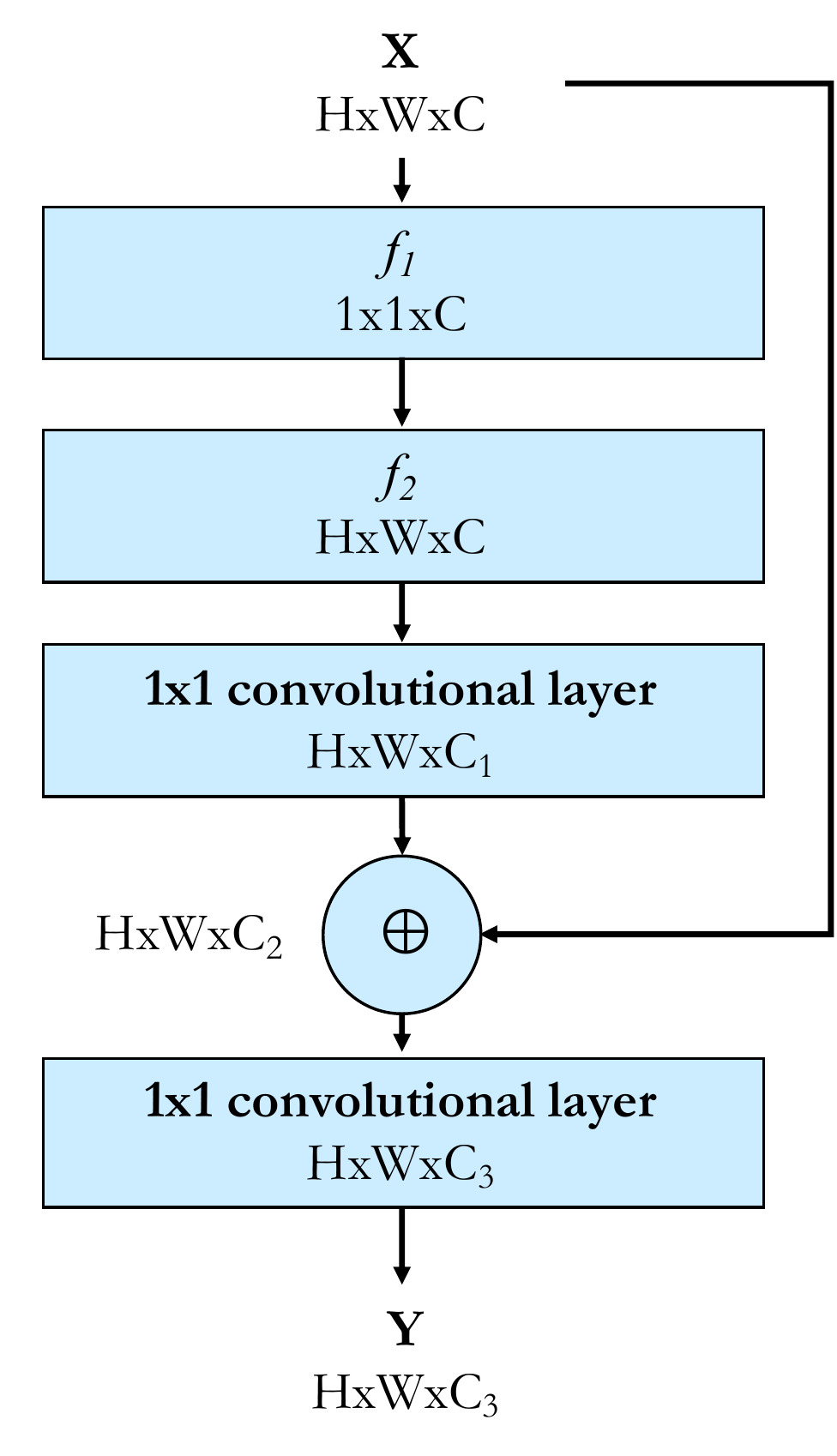}
\caption{An illustration of a global information aware (GIA) module. A GIA module consists of a stack of operations with the shapes of feature maps after each operation specified. } 
\label{fig:gia}
\end{figure}
In this paper, we propose a simple yet effective module, i.e. a global information aware (GIA) module,  to extract the global color information, which is further integrated with the pixel-level feature maps into the network for low-light imaging. 

As depicted in Fig.~\ref{fig:gia}, a GIA module consists of a stack of operations with the shapes of feature maps after each operation specified.  Particularly, given an input feature map, i.e. \textbf{X}, with a size of $\text{H}\times \text{W}\times \text{C}$, a down-sampling function $f_1(\textbf{X})$ is employed to extract the global information producing a feature map with a size of $\text{1}\times \text{1}\times \text{C}$. Then, an up-sampling function $f_2(X_1)$ is utilized to upscale the down-sampled feature map which is processed by a $1\times 1$ convlutional layer to shrink the number of channels, yielding a feature map with a size of $\text{H}\times \text{W}\times \text{C}_1$. Then a function $f_3$ is employed to combine the input feature map (encoding local information) and $\textbf{X}$ (encoding the global information) to produce an output feature map, i.e. $\textbf{Y}$, with a size of $\text{H}\times \text{W}\times \text{C}_2$. The designed GIA module is easy to be implemented and introduces negligible learnable parameters or computational cost. 

\subsection{Global Information Aware Network}

\begin{figure*}[!h]
  \centering
      \includegraphics[width=\textwidth]{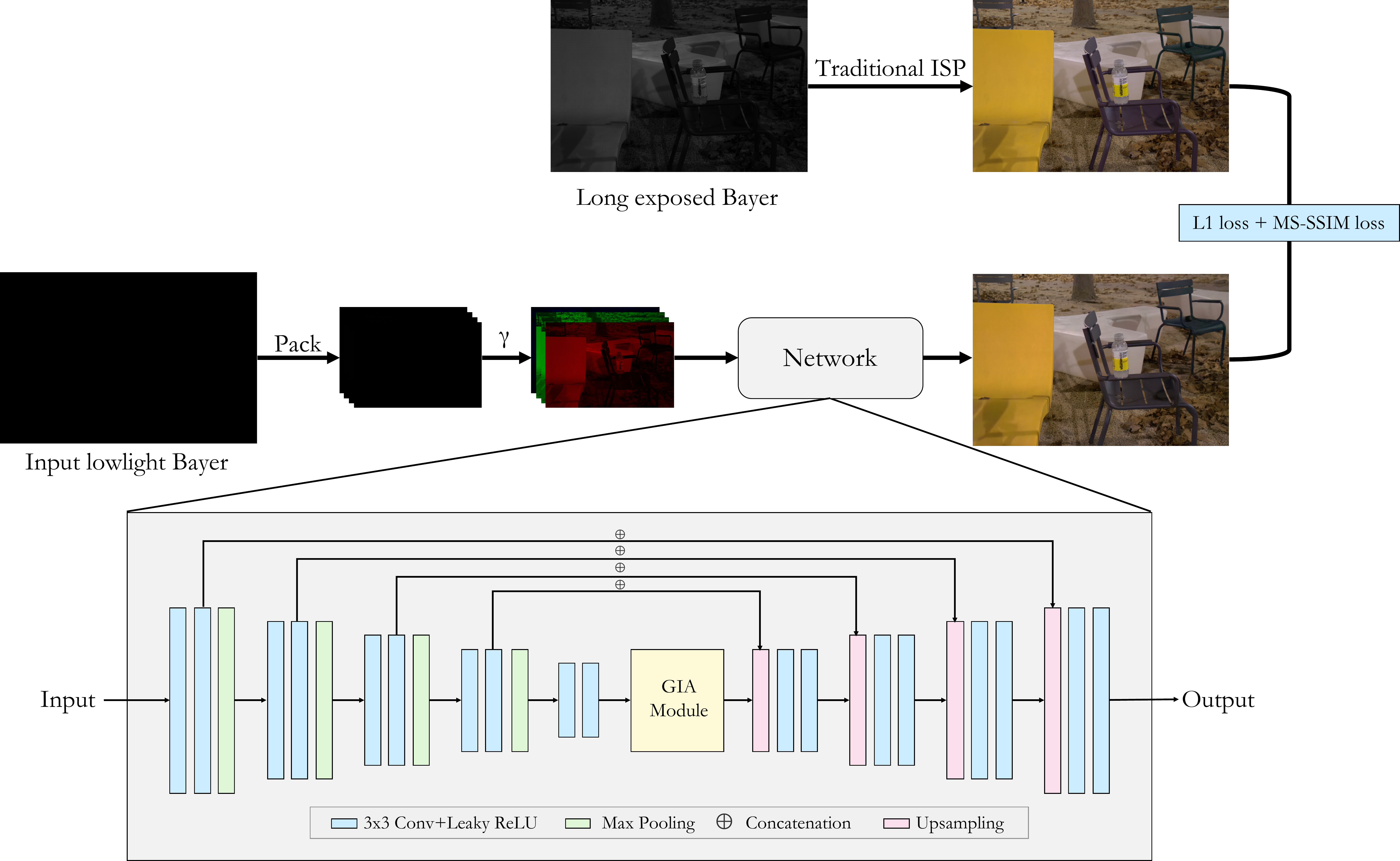}
  \caption{An illustration of the global information aware network (GIA-Net), where the base network is a vanilla U-Net with the proposed GIA module inserted into the bottleneck.}
\label{fig:network}
\end{figure*}

To illustrate the effectiveness of the proposed GIA module, we insert the GIA module into the bottleneck of a vanilla U-Net, denoted as GIA-Net, to perform low-light imaging
. Specifically, as illustrated in Fig~\ref{fig:network}, the base network is a U-Net consisting of 18 convolutional layers, represented by the blue bars. The proposed GIA module integrated in the bottleneck denoted by the yellow block. During inference, an input image firstly goes through a set of downsampling stages to extract abstract features, as well as to reduce the spatial resolution. In the bottleneck, the proposed GIA module is responsible for extracting the global information which is combined with the input feature map. Then, the feature map will go through a set of upscaling stages. In the upscaling stage, the input layer is firstly upscaled and then concatenated with the layer of the same resolution from the downsampling stage, indicated by the directed arrow which can effectively preserve the details in an image. More formally, given an input raw image, $I$, the GIA-Net is employed to learn a mapping, $\hat{I}=f(I:w)$, to produce the output RGB image, where $w$ is a set of learnable parameters of the network.

%

\subsection{Joint loss function}
We follow paper~\cite{zhao2016loss} to use a joint guidance of $\ell_1$ loss plus MS-SSIM loss. 
 The joint loss function has the following form:

\begin{equation}
\label{eq:joint}
\mathcal{L} = \gamma\mathcal{L}^{\ell_1}(I, \hat{I}) + (1 - \gamma) \mathcal{L}^{MS-SSIM}(I, \hat{I})
\end{equation}
where $\gamma\in [0,1]$ is the weight to balance the two terms.

\subsubsection*{Pixel level constraint:} The $\ell_1$ loss function calculates the difference between the ground truth image and long exposure image with the output produced by the proposed network with the corresponding short-exposed image as input. The $\ell_1$ loss function is defined as follows:

\begin{equation}
\mathcal{L}^{\ell_1} = \frac{1}{N}\sum\limits_{p=1}^N(I_p, \hat{I}_p)
\end{equation}
where $p$ is the pixel location and N gives the total number of pixels. 

\subsubsection*{Structural similarity constraint:} Although $\ell_1$ loss is widely used for image reconstruction and has been proven effective, it is reported to produce blurry results. In this work, the multiscale structural similarity index (MS-SSIM) is widely used in measuring the structural similarities of two images. SSIM is a perception-based metric which captures the similarities in structural information (i.e. pixels spatially close to each other are highly correlated), as well as the illuminance and contrast information. MS-SSIM is an extension of SSIM onto multi-scale domain. The MS-SSIM is defined as follows:
\begin{equation}
\label{eq:msssim}
\text{MS-SSIM(i)} = l_M^\alpha(i)\cdot\prod\limits_{j=1}^Mcs_j^{\beta_j}(i)
\end{equation}
where $l(i)$ and $cs(i)$ are the luminance and the product of contrast and structural difference terms at pixel $i$, respectively, which are defined as follows:
\begin{equation}
l(i) = \frac{2\mu_x\mu_y+Const_1}{\mu_x^2+\mu_y^2+Const_1}
\end{equation}
\begin{equation}
cs(i) = \frac{2\sigma_{xy}+Const_2}{\sigma_x^2+\sigma_y^2+Const_2}
\end{equation}
where $(x,y)$ gives the coordinate of pixel $i$; $\mu_x$ and $\mu_y$, $\sigma_x$ and $\sigma_y$, and $\sigma_{xy}$ are the means, standard deviations, and covariance between image $x$ and $y$, respectively, calculated using a Gaussian filter, $G_g$, with zero mean and a standard deviation $\sigma_g$; $M$ is the number of levels to perform SSIM; and $\alpha$ and $\beta_j$ for $j={i,...,M}$ are set to 1. $Const_1$ and $Const_2$ are small constant numbers~\cite{wang2003multiscale}.

The MS-SSIM is a scalar between 0 and 1, the larger the better. Thus, the final loss function used to optimize the network is given as follows:
\begin{equation}
\mathcal{L}^{MS-SSIM} = 1 - \text{MS-SSIM}
\end{equation}


\begin{table*}
\centering
\caption{Quantitative comparison between the proposed GIA-Net and the state-of-the-art methods in terms of PSNR (higher is better), SSIM (higher is better), PieAPP (lower is better), and LPIPS (lower is better). The numbers are obtained by taking the average on Sony and Fuji subsets respectively. $^*$For SID-Net, we retrained the networks using the code provided by the author of SID-Net and report the yielded numbers. Note that the numbers in the original paper is given in the parenthesis.  }
\begin{tabular}{c || c c c c | c c c c}
& \multicolumn{4}{c|}{\textbf{Sony}}  & \multicolumn{4}{c}{\textbf{Fuji}}\\\cline{2-9}
 & PSNR & SSIM & PieAPP & LPIPS & PSNR & SSIM & PieAPP & LPIPS \\\hline\hline
DnCNN~\cite{zhang2017beyond} 			& 27.79
								& 0.738
								& 1.678
								& 0.538
								& 26.23
								& 0.687
								& 1.935
								& 0.583 \\
RID-Net~\cite{anwar2019real} 			& 28.51	
								& 0.755
								& 1.577
								& 0.459
								& 26.75
								& 0.694
								& 1.915
								& 0.578\\
\multirow{2}{*}{SID-Net~\cite{chen2018learning}$^*$ } 					& 28.52 
								& 0.786 
								& 1.532 
								& 0.420 
								& 26.71 
								& 0.707 
								& 1.902 
								& 0.562 \\
								& (28.88) & (0.787)&  & & (26.61) & (0.680) & & \\
SGN~\cite{gu2019self} 			& 29.06	
								& -
								& -
								& -
								& 27.41
								& -
								& -
								& -\\
SE-UNet~\cite{hu2018squeeze} 			& 29.36	
								& 0.768
								& 1.542
								& 0.433
								& 27.78
								& 0.708
								& 1.787
								& 0.533\\
Zamir et al.~\cite{zamir2019learning} 							& 29.43	
								& -
								& 1.511
								& 0.443
								& 27.63
								& -
								&  1.763
								&  \textbf{0.476}\\
GIA-Net  			& \textbf{29.72}
								& \textbf{0.795}
								& \textbf{1.425}
								& \textbf{0.404}
								& \textbf{28.15}
								& \textbf{0.722	}		
								& \textbf{1.739 }
								& 0.519
\end{tabular}
\label{tab:results}
\end{table*}

\section{Experimental Results}

\subsection{Database}
To enable the development of low-light imaging approaches with real-world images, Chen et al.~\cite{chen2018learning} constructed a large scale dataset, i.e. See-in-the-Dark (SID) dataset. Specifically, two subsets were collected using two different sensors, i.e. Sony $\alpha$7S II with a Bayer color filter array with a resolution of $4240\times 2832$, and a Fuji X-T2 with an X-Trans CFA with a resolution of $6000\times 4000$. There are 5,094 short-exposure RAW input images with corresponding long-exposure reference images collected under both indoor and outdoor scenarios containing only static objects. The images were collected under an environment of 0.2 to 5 lux and 0.03 to 0.3 lux for outdoor and indoor  scenes, respectively. The short exposure images were taken with an exposure time of 1/30, 1/25 or 1/10 seconds and the long exposure images were taken with an exposure time of 10 seconds.

To the best of our knowledge, the SID dataset is the first and only dataset available to develop data-driven digital image processing solutions under extreme low-light conditions. Thus, in this work, we trained and evaluated our proposed method on the SID dataset~\cite{chen2018learning}.
\subsection{Implementation Details}

\noindent\textbf{Preprocessing }
There are two subsets in SID dataset constructed using two different sensors, respectively. The raw images are packed into 4 channels for Sony images with a Bayer filter array, and into 9 channels for Fuji images with an X-Trans filter array. A camera-specific black level is subtracted from the packed images. The result is then normalized into [0,1]. The normalized signal is multiplied with an amplification factor to match the brightness of its corresponding long exposure image, which is employed as input to the network. 

\noindent\textbf{Training } 
We trained two separate networks for the two subsets. For fair comparison, the base U-Net adopted the same architecture in SID-Net. Each network takes a short-exposed image preprocessed as mentioned above and yields an output image. The joint loss function, i.e. Eq.~\ref{eq:joint}, between the output and the corresponding long-exposed image, is used to guide the training process. $\gamma$ in Eq.~\ref{eq:joint} is set to 0.84 following the settings in~\cite{zhao2016loss}. Adam is employed with an initial learning rate of 0.1 for 2,000 epochs. The learning rate is decayed by a factor of 0.1 and used to train the network for another 2,000 epochs. For the GIA module, $f_1$ is global pooling, $f_2$ is bilinear interpolation, and $f_3$ is concatenation in our experiments. 


\noindent\textbf{Data Augmentation } 
Following the settings in SID~\cite{chen2018learning}, we randomly crop a patch with random flipping and transpose as the input for training the network. Moreover, to help the GIA-Net to better capture the global information from inputs with different spatial resolutions, different from the practice in ~\cite{chen2018learning,zamir2019learning} using patches of the same size, we propose to cropped patches with different sizes for training the network. Specifically, for each iteration, we randomly crop a patch with a size of $(a\times b)\times (a\times b)$ as input, where $a=32$ and $b\in[16,32]$.

\subsection{Quantitative Results}
In this work, we compare our proposed approach with several state-of-the-art methods, including DnCNN~\cite{zhang2017beyond}, RID-Net~\cite{anwar2019real}, SID-Net~\cite{chen2018learning}, SGN~\cite{gu2019self}, SE-UNet, and Zamir's method~\cite{zamir2019learning}. Note that, SE-UNet is constructed by inserting an squeeze-and-excitation (SE) module~\cite{hu2018squeeze} into the bottleneck of the same U-Net we employed. The results for RID-Net~\cite{anwar2019real}, and DnCNN~\cite{zhang2017beyond} are generated by retraining the models on SID dataset.

\begin{figure}
\centering
\includegraphics[width=\textwidth]{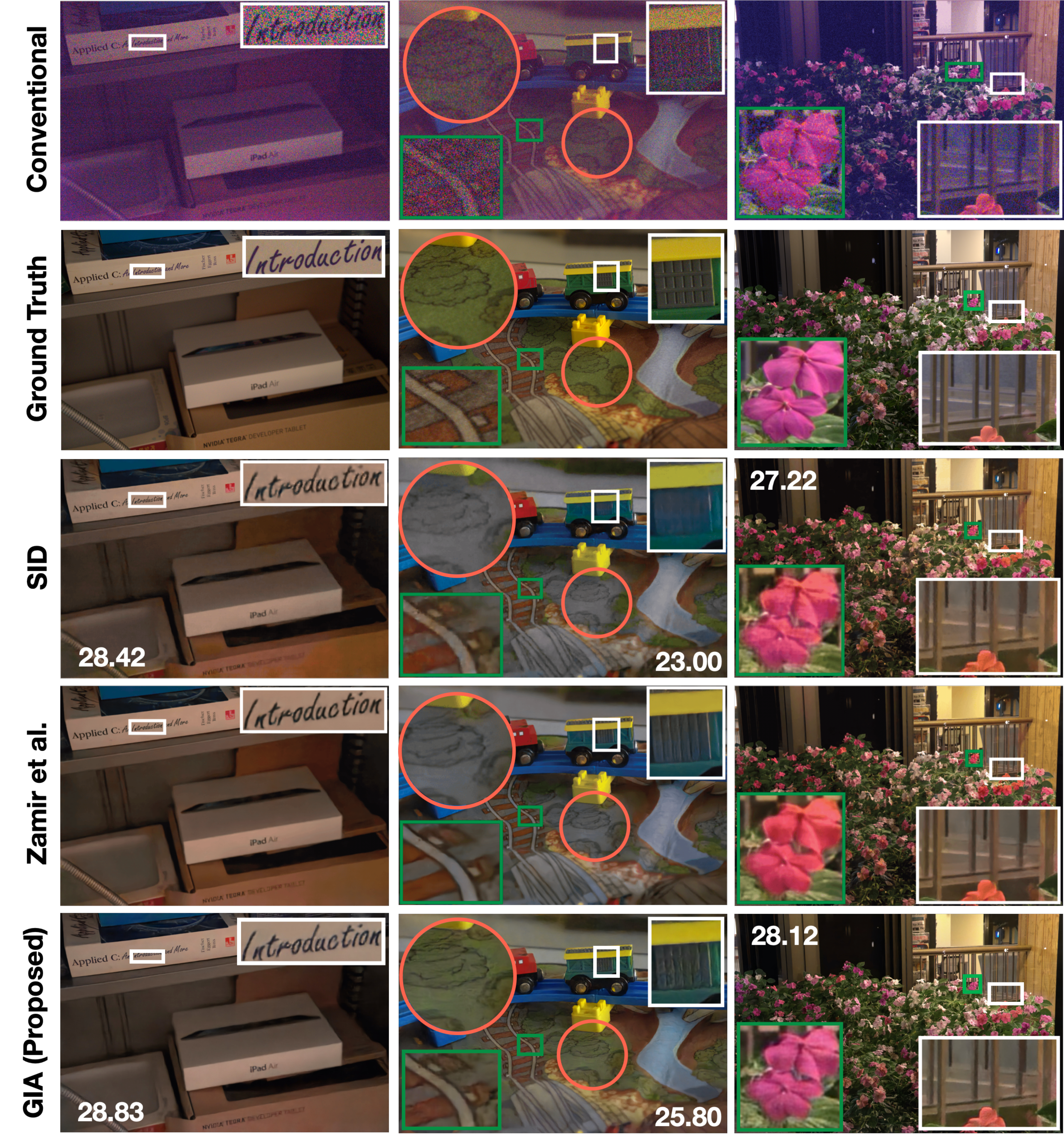}
\caption{Quantitative comparison with state-of-the-art methods. Rows 1-5 give the results generated by conventional pipeline, the ground truth, SID-Net~\cite{chen2018learning}, Zamir's approach~\cite{zamir2019learning} and the proposed GIA-Net.}
\label{fig:comparisonwithstoa}
\end{figure}

We evaluate our method with the widely used PSNR and SSIM~\cite{wang2004image} following~\cite{chen2018learning}. In addition, two recently proposed learning-based metrics, i.e. PieAPP~\cite{prashnani2018pieapp} and LPIPS~\cite{zhang2018unreasonable}, which are designed to measure the perceptual similarities between an image pair, are adopted to demonstrate the effectiveness of the proposed method. The quantitative results are obtained by taking the average of the metrics on all the testing images in Sony and Fuji subset in SID dataset, which are given in Table~\ref{tab:results}. Since the PieAPP and LPIPS values are not given in~\cite{chen2018learning}, we retrained the model using the code provided by the authors and reported the results in the table. Note that the numbers reported in the original paper are given in parenthesis. The SSIM values are omitted for Zamir's method since they are not provided in the original paper. For SGN~\cite{gu2019self}, since only PSNR values are given in the original paper, we list them in the table for comparison.

On the Fuji subset, the proposed GIA-Net outperforms all the methods in comparison for all the metrics except the LPIPS compared with Zamir's method. The reason is because LPIPS and Zamir et al. both employed the same pretrained VGG network to calculate the perceptual loss. On the Sony dataset, our method outperforms the state-of-the-art methods significantly in terms of all the metrics employed including the LPIPS compared with Zamir's approach, which have demonstrated the effectiveness of the proposed GIA-Net for low-light imaging with exceptional perceptual quality. Note that, although the SE-module is operation in a similar way with GIA-module by applying global average pooling, it is not as effective as the GIA module in terms of integrating the global information into the network for low-light image enhancement.

\subsection{Qualitative Results}
Fig.~\ref{fig:comparisonwithstoa} gives the quantitative comparison with the state-of-the-art methods, where rows 1-5 provide the images generated by the conventional image pipeline, the ground truth, SID-Net, Zamir's method, and the proposed GIA-Net, respectively. Since the code or the result images were not released in~\cite{zamir2019learning}, to perform qualitative comparison with the state-of-the-arts, we took the Fig. 5 in~\cite{zamir2019learning} and extended it by adding the ground truth images and the results produced by our proposed GIA-Net.  The qualitative results measured by PSNR are shown on the images (the values are omitted for Zamir's method since they are not provided). Note that the zoomed-in areas are directly adopted from~\cite{zamir2019learning} except the region highlighted by the orange circle in the middle column, where GIA-Net achieves better performance compared with SID and comparable performance compared with Zamir's approach in terms of detail reconstruction in the areas highlighted by the white and green rectangles. More importantly, the proposed GIA-Net produces better color representation compared with the other two approaches. For example, the area highlighted by the orange circle in the second column is green, as shown in the groundtruth image. However, SID-Net and Zamir's method failed to restore the green color, while the proposed GIA-Net successfully captured the green color thanks to the integration of the global color information. In addition, in the third column, the flowers highlighted in the image generated by GIA-Net has the same color with those in the ground truth image, which has further demonstrated the effectiveness of introducing  global information into the network for low-light imaging.

\subsection{Ablation Study}

\noindent\textbf{Importance of global information }

To validate the importance of extracting and exploiting the global information for low-light imaging, we compare the performance of the proposed GIA-Net with the following models: (1) the original SID-Net; (2) a network with all the convolutional layers replaced by dilated convolutional layers with dilated rate as 2, denoted as SID-dilated, which has larger receptive field size than SID-Net while the local information is not well utilized; (3) a network constructed by replacing all the convolutional layers with the SW block, denoted as SW-Net, which utilizes both local information and wider contextual information as receptive field size gradually increases; (4)  a vanilla U-Net trained using $\ell_1$ loss with GIA module inserted into the bottleneck, denoted as GIA-$\ell_1$. For SW-Net, we set $C_l=C_g=C_{in}/2$ in SW block using a dilate rate of 2 as illustrated in Fig.~\ref{fig:seewider}. Note that SID-dilated, SW-Net and GIA-$\ell_1$ are proposed in this work. 
The experimental results are reported in Table~\ref{tab:seewider}. The proposed GIA-$\ell_1$ achieves the best performance thanks to the utilization of global information extracted by the integrated GIA module. More importantly, on Fuji subset, although SID-dilated fails to outperform SID-Net in terms of PSNR  and SSIM, it achieves better performance measured by PieAPP and LPIPS, which illustrates that the global information is crucial for generating images with good perceptual quality. 

\begin{table*}
\centering
\caption{Performance comparison using SID-Net~\cite{chen2018learning},SID-dilated, SW-Net, and GIA-$\ell_1$.}
\begin{tabular}{c || c c c c | c c c c}
 & \multicolumn{4}{c|}{\textbf{Sony}}  & \multicolumn{4}{c}{\textbf{Fuji}}\\\cline{2-9}
 & PSNR & SSIM & PieAPP & LPIPS & PSNR & SSIM & PieAPP & LPIPS \\\hline\hline
SID-Net~\cite{chen2018learning}  					& 28.52
								& 0.786
								& 1.532 
								& 0.420 
								& 26.71
								& 0.707
								& 1.902 
								& 0.562 \\
SID-dilated				& 28.62	
								& 0.780
								& 1.583
								& 0.417
								& 26.61
								& 0.695
								&  1.759
								&  \textbf{0.536}\\
SW-Net				& 28.89	
								& 0.787
								& 1.508
								& 0.417
								& 27.05
								& 0.708
								&  1.872
								&  0.548\\
GIA-$\ell_1$  				& \textbf{29.45}
								& \textbf{0.790}
								& \textbf{1.449}
								& \textbf{0.410}
								& \textbf{27.48}
								& \textbf{0.711}		
								& \textbf{1.819 }
								& 0.548
\end{tabular}
\label{tab:seewider}
\end{table*}

\begin{figure}
\centering
\includegraphics[width=0.65\textwidth]{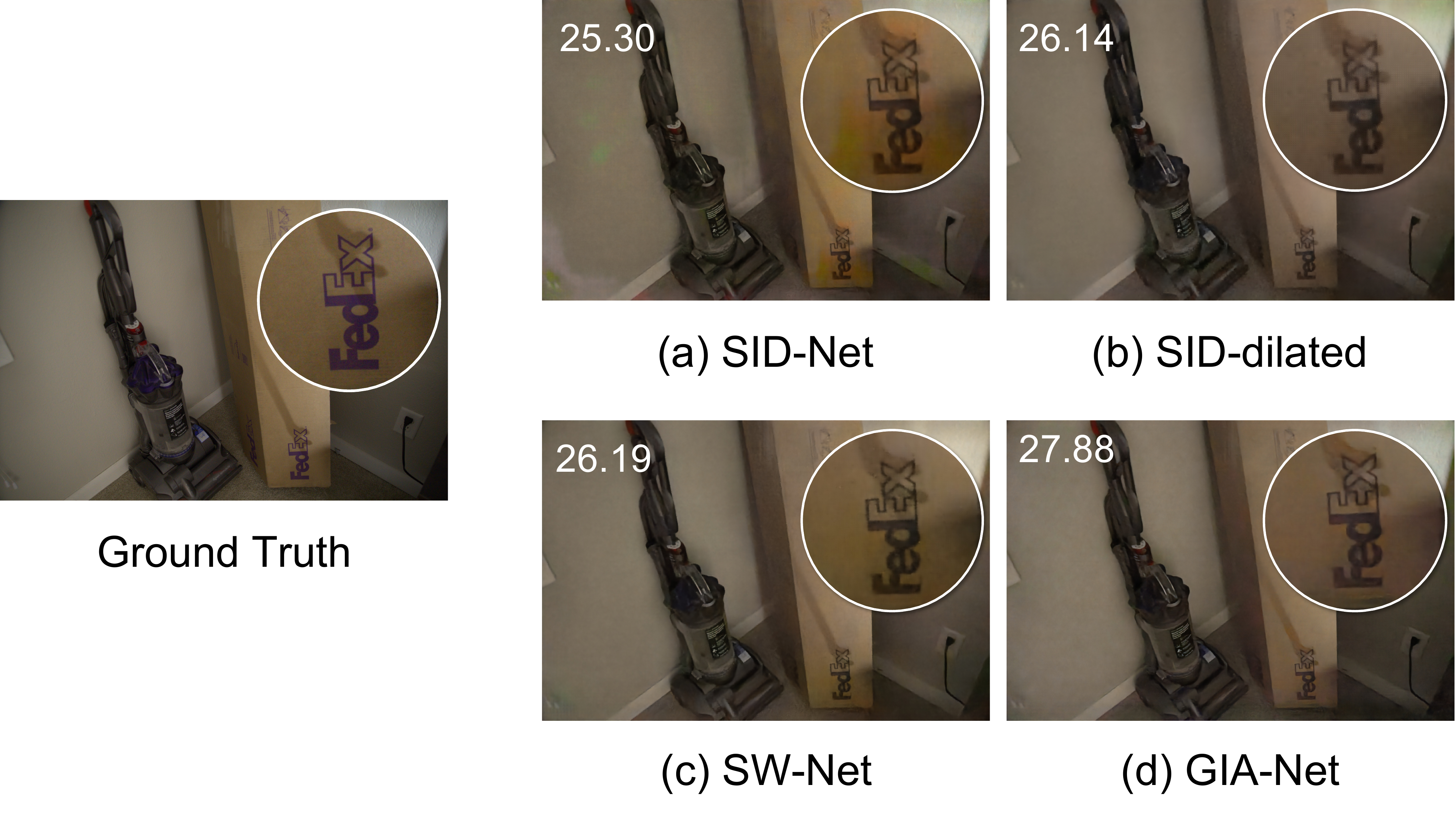}
\caption{A comparison of the results produced by different models.
}
\label{fig:widercomp}
\end{figure}

Fig~\ref{fig:widercomp} gives some qualitative results, where (a), (b), (c), (d) give results produced by SID-Net, SID-dilated, SW-Net, and the GIA-Net with only $\ell_1$ loss and the PSNR values are reported on the images. Severe color artifacts can be spotted on the image produced by SID-Net due to the lack of global color information. Although the perceptual quality is much better in the image generated by SID-dilated, it becomes blurry because the dilated convolutions do not fully use all the pixels in the local neighborhood. In the image yielded by SW-Net, both the color and details are somehow well restored. However, the dilation rate and the $C_l$, $C_g$ require extensive experiments to tune to achieve optimal results. The image produced by the proposed GIA-Net gives good perceptual quality with fine details, demonstrating the effectiveness of the proposed GIA module to extract and integrate the global information into the network for low-light imaging.

%

\noindent\textbf{Model Analysis }
To better understand the proposed model, we conducted controlled experiments to evaluate how much each component contribute to the final performance. We compare the models with or without $\ell_1$ loss, MS-SSIM, GIA, and data augmentation. Besides, to validate the performance does not come from the depth increase, i.e. two more convolutional layers are introduced in to the network by GIA module, we also conduct an experiment using a network with two convolutional layers added to the bottleneck of SID-Net. The results can be found in Table~\ref{tab:modelanalysis}.

\begin{table*}
\centering
\caption{Performance comparison between models with different component choices on the SID datasets. }
\begin{tabular}{c | c  || c c c c c | c c c c}
\multicolumn{2}{c||}{} & \multicolumn{5}{c|}{SID-Net based Models} & \multicolumn{4}{c}{GIA-Net based Models}  \\\hline\hline
\multicolumn{2}{c||}{No.} & 1 & 2 & 3 & 4 & 5 & 6 & 7 & 8 & 9 \\\hline
\multicolumn{2}{c||}{$\ell_1$ Loss}				& $\surd$ 
								& $\surd$ 
								& $\surd$ 
								& $\surd$ 
								& $\surd$ 
								& $\surd$ 
								& $\surd$ 
								& $\surd$ 
								& $\surd$ \\
\multicolumn{2}{c||}{GIA module}						&  			
								&  			
								&  
								& 
								& 
								& $\surd$ 
								&$\surd$
								& $\surd$ 
								& $\surd$ \\
\multicolumn{2}{c||}{MS-SSIM Loss} 			&  
								& 			
								& $\surd$
								& 
								& $\surd$
								& 
								& $\surd$ 		
								& 
								& $\surd$ \\
\multicolumn{2}{c||}{Data Augmentation}  	&  			
								&  			
								&  
								& $\surd$
								& $\surd$ 
								& 
								&	 		
								& 	$\surd$ 
								& $\surd$ \\
\multicolumn{2}{c||}{Additional 2 Conv. Layers}  	&  		
								&  $\surd$	
								&  
								& 
								& 
								& 
								&	 		
								& 	
								&  \\\hline\hline
\parbox[t]{4mm}{\multirow{4}{*}{\rotatebox[origin=c]{90}{Sony}}} &	PSNR		& 28.52
								& 28.76 	
								& 28.62 
								& 28.57	
								& 28.73
								& 29.45
								& 29.62
								& 	29.65	
								&  \textbf{29.72} \\
								& SSIM & 0.786 	
								& 0.788 
								& 0.790 
								& 0.787	
								& 0.790
								& 0.790
								& 0.793
								& 	0.791
								&  \textbf{0.795} \\
								& PieAPP & 1.532 	
								& 1.513 
								& 1.507 
								& 1.539	
								& 1.494
								& 1.449
								& 1.445
								& 	1.425
								&  \textbf{1.425} \\
								& LPIPS & 0.420 	
								& 0.416 
								& 0.415 
								& 0.419	
								& 0.414
								& 0.410
								& 0.409
								& 	0.408
								&  \textbf{0.404} \\\hline
\parbox[t]{4mm}{\multirow{4}{*}{\rotatebox[origin=c]{90}{Fuji}}} 	& PSNR & 26.71 	
								& 26.70 	
								& 26.69 	
								& 26.64
								& 26.76
								& 27.48
								& 27.74
								& 28.06	
								& \textbf{28.15} \\
								& SSIM & 0.707	
								& 0.706	
								& 0.711	
								& 0.706
								& 0.712
								& 0.711
								& 0.717
								& 	0.713
								& \textbf{0.722} \\
								& PieAPP & 1.902	
								& 1.853	
								& 1.833	
								& 1.882
								& 1.877
								& 1.819
								& 1.757
								& 	1.765
								& \textbf{1.739} \\
								& LPIPS & 0.562	
								& 0.550 	
								& 0.527 	
								& 0.551
								& 0.535
								& 0.548
								& 0.531
								& 0.540
								& \textbf{0.519} \\
\end{tabular}
\label{tab:modelanalysis}
\end{table*}

\textbf{Depth increase} does bring improvements to the final performance, e.g. 28.76 (model 2) v.s. 28.52 (model 1) measured in PSNR on Sony subset. However, the major performance gain is from the integration of the proposed GIA moduel as indicated by the comparison between model 2 and model 6, e.g. 28.76 (model 2) v.s. 29.45 (model 6) measured in PSNR on Sony subset.

\textbf{Importance of GIA module} is emphasized by the comparison between the models without GIA modules (i.e. model 1, 3, 4, 5) with their counterparts using GIA modules (i.e. model 6, 7, 8, 9). More importantly, without using GIA module in the  network, all the other techniques, i.e. using MS-SSIM loss (model 3), data augmentation (model 4), and both (model 5), yield similar performance with the original SID-Net. In contrast, the models with GIA modules integrated can benefit from using MS-SSIM loss and data augmentation. 

\noindent\textbf{Computational cost introduced by the GIA module}
Table~\ref{tab:computational} gives the comparison of the number of parameters and FLOPs of the proposed GIA-Net, relative to the SID-Net~\cite{chen2018learning} processing an image from Sony subset with a resolution of $4240\times 2832$. The proposed GIA-Net achieves much better performance in terms of all the metrics employed than SID-Net (e.g. according to Table~\ref{tab:results}, 29.72 v.s. 28.52 measured in PSNR on Sony subset) with negligible extra computational cost ($0.008\times$ increase in FLOPs). 

\begin{table}
\centering
\caption{Comparison of numbers of params and FLOPs between SID-Net~\cite{chen2018learning} and GIA-Net processing an input image from Sony dataset with a resolution of $4240\times 2832$. The numbers of parameters and FLOPs are relative to the SID-Net (7.76M and 1112.92B).}
\begin{tabular}{c | c  c}
Model & Params & FLOPs \\\hline\hline
SID-Net & $1\times$ & $1\times$ \\\hline
GIA-Net & $1.07\times$ & $1.008\times$
\end{tabular}
\label{tab:computational}
\end{table}

\section{Conclusion}
Taking images with good perceptual quality is challenging due to low SNR under extremely low-light conditions. Most recently, deep U-Nets have show promising results on low-light imaging. However, vanilla U-Nets suffer from color distortion due to the lack of global information. In this paper, we propose a GIA module which can be inserted into a vanilla U-Net to extract and integrate global information into the network to improve the perceptual quality of the generated image for low-light imaging. The experimental results on a public dataset have demonstrated the effectiveness of the proposed approach. 
In the future, we plan to explore the possibilities of applying it in a multi-scale fashion to better extract the color information to further improve the performance for low-light imaging. Also, we would like to explore the possibilities to apply the proposed GIA module to other computer vision applications, such as image segmentation and image deblurring. 

\bibliographystyle{splncs04}
\bibliography{abbrev,literature-denoising, }

\end{document}